%% file: amvnet.tex
\documentclass[final]{cvpr}

\usepackage{times}
\usepackage{epsfig}
\usepackage{graphicx}
\usepackage{amsmath}
\usepackage{amssymb}
\usepackage{multirow} % Cells that span multiple rows in a table
\usepackage{adjustbox}
\usepackage{array}
\usepackage{booktabs}
\usepackage{subcaption}
\usepackage{pifont}% http://ctan.org/pkg/pifont
\newcommand{\cmark}{\ding{51}}%
\usepackage{enumitem}

\newcommand{\mypar}[1]{\vspace{-5mm}\paragraph{#1}} % Customized paragraph without vspace
\usepackage[normalem]{ulem} % for striking out text with \sout
\newcommand{\squeeze}{\vspace{-1mm}} % To add in various places to quickly adjust whitespaces
\usepackage{tabularx}
\newcolumntype{C}[1]{>{\centering\let\newline\\\arraybackslash\hspace{0pt}}m{#1}} % column-type in table
\newcolumntype{L}[1]{>{\let\newline\\\arraybackslash\hspace{0pt}}m{#1}} % column-type in table
\usepackage{siunitx} % For degrees symbol
\usepackage{float} % For figure placement

\newcolumntype{R}[2]{%
    >{\adjustbox{angle=#1,lap=\width-(#2)}\bgroup}%
    l%
    <{\egroup}%
}
\newcommand*\rot{\multicolumn{1}{R{45}{1em}}}% no optional argument here, please!

\usepackage[normalem]{ulem} % for striking out text with \sout

\usepackage[pagebackref=true,breaklinks=true,colorlinks,bookmarks=false]{hyperref}

\def\cvprPaperID{} % *** Enter the CVPR Paper ID here
\def\confYear{}

\begin{document}

%%%%%%%%% TITLE
\title{AMVNet: Assertion-based Multi-View Fusion Network for LiDAR Semantic Segmentation}
\author{
Venice Erin Liong$^{\dagger}$ \and
Thi Ngoc Tho Nguyen$^{\dagger}$\thanks{Work done during her internship at Motional}
\and Sergi Widjaja$^{\dagger}$
\and Dhananjai Sharma$^{\dagger}$
\and Zhuang Jie Chong$^{\dagger}$ \\
$^{\dagger}$Motional \hspace{5mm} $^*$Nanyang Technological University\\
\tt\small{ $^\dagger$\{venice.liong, sergi.widjaja, dhananjai.sharma, demian.chong\}@motional.com} \\
\tt\small $^*$nguyenth003@e.ntu.edu.sg
}

\maketitle
%%%%%%%%% ABSTRACT
\begin{abstract}
In this paper, we present an Assertion-based Multi-View Fusion network (AMVNet) for LiDAR semantic segmentation which aggregates the semantic features of individual projection-based networks using late fusion.
Given class scores from different projection-based networks, we perform assertion-guided point sampling on score disagreements and pass a set of point-level features for each sampled point to a simple point head which refines the predictions.
This modular-and-hierarchical late fusion approach provides the flexibility of having two independent networks with a minor overhead from a light-weight network.
Such approaches are desirable for robotic systems, e.g. autonomous vehicles, for which the computational and memory resources are often limited. 
Extensive experiments show that AMVNet achieves state-of-the-art results in both the SemanticKITTI and nuScenes benchmark datasets and that our approach outperforms the baseline method of combining the class scores of the projection-based networks.
\end{abstract}

%%%%%%%%% BODY TEXT
\input{sections/introduction.tex}
\input{sections/relatedwork.tex}
\input{sections/method.tex}

\input{sections/experiments.tex}
\input{sections/conclusion.tex}

{\small
\bibliographystyle{ieee_fullname}
\bibliography{amvnet}
}
\end{document}

%% file: sections/introduction.tex
% !TEX root = ../amvf.tex
\vspace{-5mm}
\squeeze
\section{Introduction}
\label{sec:introduction}
\squeeze
\vspace{-2mm}
% Importance of LidarSeg
Point cloud semantic segmentation is a critical task for autonomous systems.
Particularly, for autonomous vehicles (AVs), this task provides useful semantic information to build crisp, high-definition maps from LiDAR point clouds.
It also helps with identifying and locating dynamic objects and drivable surfaces for perception modules. 
This results in better vehicle maneuvering and path planning.
\begin{figure}
\begin{center}
\includegraphics[width = 0.5\textwidth]{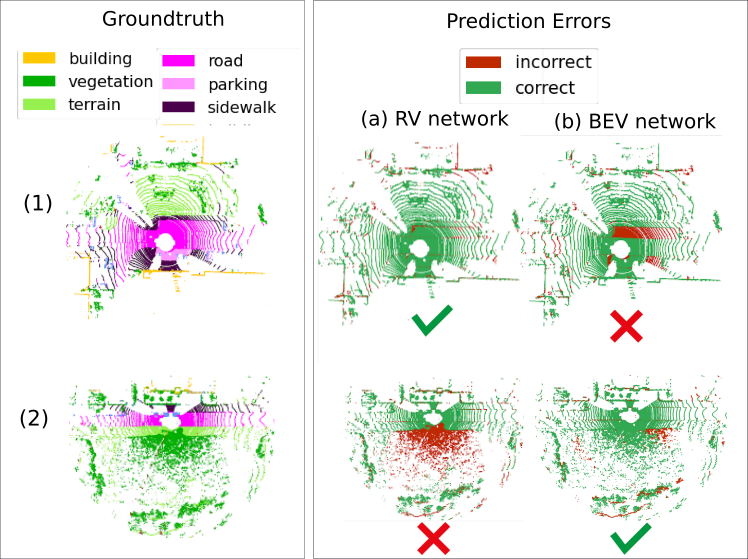}
\end{center}
\vspace{-6mm}
\caption{Point cloud prediction errors visualized in top view after evaluating vanilla  (a) Range view (RV) and (b) Bird's-eye view (BEV) networks given LiDAR samples (1) and (2). The RV network performs better in (1), while BEV performs better in (2).
This suggests each network can provide unique and useful \mbox{semantic} information which motivates our proposed late fusion approach.
\label{fig:errors}}
\end{figure}

Recently, more AV datasets with point-level annotations have emerged, such as SemanticKITTI~\cite{Semantickitti} and nuScenes~\cite{Nuscenes}, and have played a significant role in advancing the field of point cloud semantic segmentation in both academia and industry alike.
The proposed solutions so far can work directly with raw 3D point clouds (point-based)~\cite{pointnet, RandLA-Net}, projected point clouds such as range view (RV)~\cite{Rangenet++, Salsanext, Kprnet} or bird's-eye view (BEV)~\cite{Polarnet}, or combinations of these projected views~\cite{Tornadonet}.
Based on recent works~\cite{Tornadonet, pcreview}, it has been shown that projection-based methods are faster and more accurate than point-based methods for AV.

RV methods contribute to a majority of state-of-the-art results in the SemanticKITTI leader board\footnote{http://semantic-kitti.org/tasks.html}~\cite{Salsanext, Kprnet, Squeezesegv3}.
RV methods leverage on a compact representation of sparse point clouds in the form of 2D pseudo-images which can be efficiently segmented using image-based semantic segmentation networks~\cite{Unet, deeplab}.
This can then be further enhanced by using tricks similar to image-based segmentation methods. 
For example, we have employed Recurrent Neural Networks (RNNs) to account for the spatial relationship of objects in the images~\cite{reseg}.
RV also presents some advantages where small, vertically-oriented classes such as \emph{pedestrians}, \emph{bicyclists}, and \emph{pole} have an adequate structural representation.
However, as the number of LiDAR points increases, the RV images would have overlapping 3D point projections in one pixel which makes it less representative.
A BEV approach will not have this difficulty as the points are projected from the top view,  where we extract the pillar representations to form the 2D pseudo-images.
Hence, the BEV projection is effective at discerning objects in space at different areas of the point cloud even at farther range. 
However, as noted by past works \cite{Polarnet, PointPillars}, the non-uniform and sparse nature of point clouds in BEV remains a limitation for standard convolution operations. 

\figref{fig:errors} further illustrates the subtle yet crucial differences between the output predictions of an RV and a BEV segmentation networks using two LiDAR scans taken from the SemanticKITTI dataset.
For the first LiDAR scan, the RV network performs better than the BEV network at classifying \emph{parking} and \emph{road}, wherein these object instances are at a relatively close position to the ego-vehicle.
For the second LiDAR scan, the BEV network performs better than the RV network at distinguishing \emph{vegetation} from \emph{terrain}.
These are just a few of many instances where RV performs better than BEV and vice versa. 
Furthermore, our analysis on class-wise IOU for both networks also suggests that there is no clear winner across all classes and that there is complementary information that can be extracted from both.
Motivated by these findings, we propose a late fusion approach that aims to combine the advantages of both the RV and BEV networks to achieve state-of-the-art results in point cloud semantic segmentation.

Our proposed approach is inspired by PointRend~\cite{Pointrend} for image segmentation which selects uncertain pixel areas and passes point features to a point head for better predictions.
Our method is different from PointRend's uncertainty criterion since we introduce a novel sampling strategy to select LiDAR points through a \emph{multi-view assertion}.
This assertion takes into account the class-prediction disagreements between the RV and BEV networks.
We flag the points where the networks disagree as \emph{uncertain}.
Thereafter, these uncertain points are passed to a light-weight \emph{point-head} network to obtain more robust predictions.
To summarize, our contributions are as follows:
\begin{itemize}[noitemsep,nolistsep]
	\item A multi-view fusion network for point cloud semantic segmentation that reaps the benefits of the RV and BEV networks; where we perform late fusion with an assertion-guided sampling strategy and take features of sampled points to a point head for more accurate predictions.
	\item A light-weight point head architecture for predicting class labels in 3D space, in which the input is a set of point features obtained from the network predictions and raw point cloud.
	\item An enhanced RV network which employs recurrent neural networks (RNNs), particularly, Gated Recurrent Units (GRUs), with circular padding to learn spatial relationships. 
	\item Extensive analysis to verify the effectiveness of our proposed method, and show that our method achieves competitive performance on the SemanticKITTI and nuScenes dataset.
\end{itemize}
To the best of our knowledge, this is the first work that explores uncertainty in the late fusion approach for LiDAR  semantic segmentation.
Our late fusion method is unique as it only processes uncertain points.
This makes it favorable as we are able to combine the benefits of the RV and BEV networks without significantly slowing down training and inference.
In addition, it is modular and flexible, as each component in the network can be improved independently to boost the overall performance.

%% file: sections/relatedwork.tex
% !TEX root = ../amvf.tex
%\vspace{+2mm}
\squeeze
\section{Related Work}
\label{sec:related_work}
\squeeze
We first present general work on multi-view fusion methods in point clouds for AVs.
Then, we present the current state-of-the-art for LiDAR semantic segmentation.
\squeeze
\subsection{LiDAR Multi-view Fusion Methods}
\squeeze
Deep learning models have been increasingly popular in extracting semantic information from point clouds through \textit{3D object detection} and, lately, \textit{semantic segmentation}.
While there exist point-wise methods which directly operate in 3D space, most state-of-the-art approaches take advantage of projections in both voxels, RV and BEV due to their ease of implementation, faster inference, and good performance. 
More recently, multi-view fusion approaches using these different networks have been explored and showed promising results\cite{MV3D, MVLidarNet, MVF, PillarFusion}; with most works on 3D object detection.
In the following, we discuss different fusion works for LiDAR.

MV3D~\cite{MV3D} combines features from images in camera view and LiDAR points in BEV and RV before passing them to a region-based fusion network to extract and classify bounding boxes.
MVLidarNet~\cite{MVLidarNet} performs sequential fusion similar to PointPainting~\cite{Pointpainting} where they first extract point-level class scores from the RV scan and re-project them to BEV before doing 3D object detection.
MVF~\cite{MVF} performs early-feature fusion by first obtaining point-level semantics from BEV and RV voxelization before proceeding to VoxelNet-like training.\cite{Voxelnet}.
Wang~\emph{et. al.}~\cite{PillarFusion} presents a pillar-based fusion approach, where they have a light-weight encoder-decoder network for BEV and RV to obtain richer point features, which they re-project back to the BEV to proceed with a PointPillar-like~\cite{PointPillars} anchor-free object detection and classification.

TornadoNet~\cite{Tornadonet}, which currently is one of the few fusion methods for LiDAR semantic segmentation, does end-to-end sequential feature fusion.
It learns pillar features from the BEV using a pillar-projection learning module and have these features as additional input channels for their RV network.

While most fusion methods explore early-feature fusion \cite{MVF, PillarFusion, FusionNet} and sequential fusion \cite{Tornadonet, MVLidarNet}, there is not much work on late fusion, especially in LiDAR semantic segmentation.
One possible reason is the difficulty in combining features from different views in later stages without projecting them back to the 3D view and processing them again.
This leads to increased overhead in point-level computation.
Our proposed method can alleviate this bottleneck by only selecting points that are flagged as \textit{uncertain} through disagreements of two projection-based networks.
Furthermore, a late fusion method also provides the flexibility of having two independent networks which can do parallel inference with the option to have more accurate predictions using a light-weight point head for fusion.

\squeeze
\subsection{LiDAR Semantic Segmentation}
\squeeze
Advancements on the LiDAR or point cloud semantic segmentation research are developing thanks to the increasing availability of public datasets~\cite{Semantickitti, Nuscenes}.
Currently, these methods can be grouped into the following categories: \textit{point-based}, \textit{voxel-based}, and~\textit{projection-based}.
Point-based methods, like PointNet~\cite{pointnet}, PointNet++~\cite{pointnet++} and SPG~\cite{SPG} work directly on raw 3D point clouds and can handle the unordered and unstructured nature of point clouds.
These networks show good results in smaller point clouds, typically indoors where points are generally dense and have uniform density.
But these would not be efficient in outdoor scenes where there is varying density.
Furthermore, getting the full 360$^\circ$ scan would lead to slow inference time and large memory requirement\cite{Cylinder3d}.

Voxel-based methods group points into voxels and then apply 3D convolutions for semantic segmentation. 
To make the methods efficient, these require special convolution operations to handle sparsity of the representation and/or speed-up processing. 
For example, Cylinder3D\cite{Cylinder3d} extracts voxels from 3D cylinder partitions and feeds them to a network with asymmetric residual blocks to reduce computation.
SPVNAS\cite{SPVNAS} uses a sparse point-voxel convolution (SPConv) which fuses features from a high-resolution light-weight point branch and sparse voxel-based branch.
FusionNet\cite{FusionNet} uses a mini-PointNet and a sparse convolution layer to effectively aggregate the features.

Projection-based methods project point clouds to a more structured and dense 2D representation. 
These point clouds are projected to either RV\cite{Rangenet++, Salsanext, Kprnet, Squeezeseg, Squeezesegv2, Squeezesegv3} or BEV\cite{Polarnet, Salsanet}, and are then fed to an encode-decoder network, like any image-based semantic segmentation problem to obtain semantic information.
For instance, SalsaNext\cite{Salsanext} shares several small enhancements on the encoder-decoder network (e.g. dilated convolutions, pixel-shuffle layer) and introduces uncertainty estimation for each point.
KPRNet\cite{Kprnet} presents an improved architecture for RV networks by using a ResNeXt-101 backbone with Atrous Spatial Pyramid Pooling, and applying KPconv\cite{Kpconv} in the last layer.
PolarNet\cite{Polarnet} represents the LiDAR points in BEV in a polar grid  to mitigate the imbalanced distribution of LiDAR points across space.
Instead of only performing segmentation in the pillar-view, they further segment pillars into voxels, to have a higher output resolution along the z-axis. 

While projection-based methods are generally simple and straightforward, there are, however, disadvantages of these structured methods. 
RV methods can have multiple 3D points be projected into a common pixel which can lead to label re-projection errors.
Hence, most RV networks perform an additional post-processing step like a projective nearest neighborhood search~\cite{Rangenet++}.
On the other hand, BEV networks present difficulties in highly sparse pillars and representing vertically-oriented objects.
Nevertheless, by leveraging the strengths of both, we can achieve better label predictions overall.

%% file: sections/method.tex
% !TEX root = ../amvf.tex
\begin{figure*}
\begin{center}
\includegraphics[width = 1.0\textwidth]{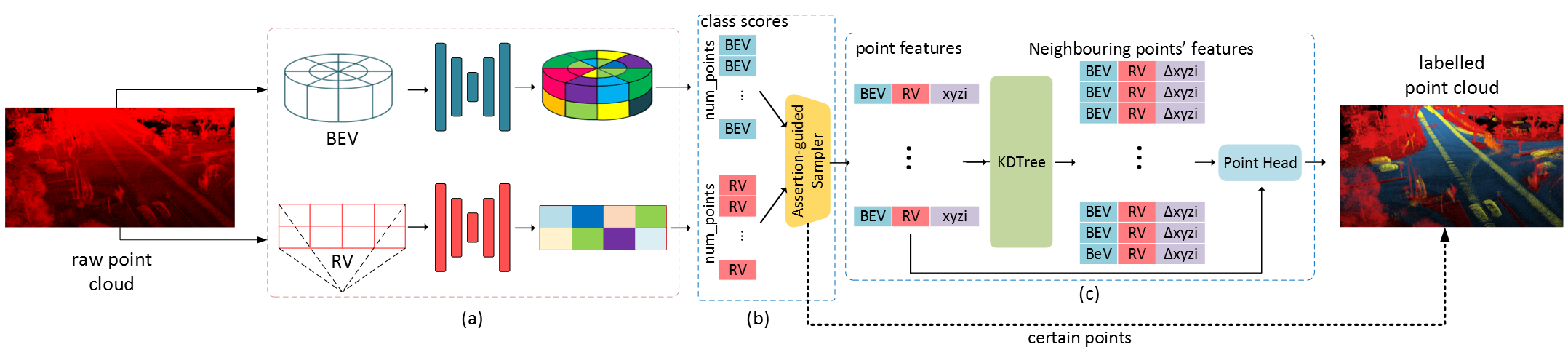}
\end{center}
\vspace{-7mm}
\caption{AMVNet overview. The AMVNet consists of three main stages: (a) Given a LiDAR scan, we project these point clouds to a structured representation and feed these to their respective multi-view networks to obtain initial point-level class predictions; (b) For each point, we categorize it to be \emph{certain} or \emph{uncertain} through an assertion-guided point sampler; (c) For each uncertain point, we extract its respective point-level features and its neighborhood points' features and feed them to the point head to obtain the final predictions. 
}
\label{fig:overall}
\end{figure*}
\vspace{-2mm}
\squeeze
\section{Assertion-based Multi-View Fusion Network (AMVNet)}
\label{sec:amvf}
\squeeze
In this section we describe the overall framework of our approach, which consists of three main parts (\figref{fig:overall}):
(1) Semantic segmentation from individual projection-based networks to obtain initial point-level class predictions; (2) Assertion-guided point sampling strategy to determine \emph{uncertain} points; (3) Point head architecture that accepts point features to obtain new predictions.

\squeeze
\vspace{-1mm}
\subsection{Multi-view Networks}
\label{sec:main_networks}
\squeeze
Given a LiDAR scan, we project the point cloud to structured representations, which are suitable inputs for any encoder-decoder network to perform semantic segmentation. From these segmentation outputs, we can re-project the class scores for each LiDAR point.
In our work, we focus on variants of RV and BEV networks as follows:
\vspace{-1mm}
\mypar{Range view network.}
We project a 360$^\circ$ LiDAR scan to a cylindrical or a spherical RV pseudo-image as input to the network where each pixel is composed of the 3D coordinates (x, y, z), intensity, range, and binary mask. 
The binary mask indicates whether or not a LiDAR point occupies that pixel.
In the case of multiple points projected onto the same pixel, we choose the point which is closest to the LiDAR sensor to represent that pixel.
We also want to have a reasonably high 3D point coverage to reduce errors due to re-projections.
Hence, the height of our input depends on the number of rings available for the sensors, and the width depends on the horizontal angular resolution. 

For the sake of clarity, we have kept our network architecture simple due to ease of implementation.
We design a Fully Convolutional Network~\cite{FCN} where we use a ResNet-like\cite{Resnet} backbone with skip connections and strided convolutions. 
Different from image-based segmentation networks~\cite{FCN, Unet, deeplab}, we only downsample the height dimension at later feature maps to accommodate the smaller height of RV images as relative to its width.
For training, we use a combination of cross-entropy (CE) and Dice loss~\cite{Diceloss}.

\begin{figure}
\begin{center}
\includegraphics[width = 0.5\textwidth]{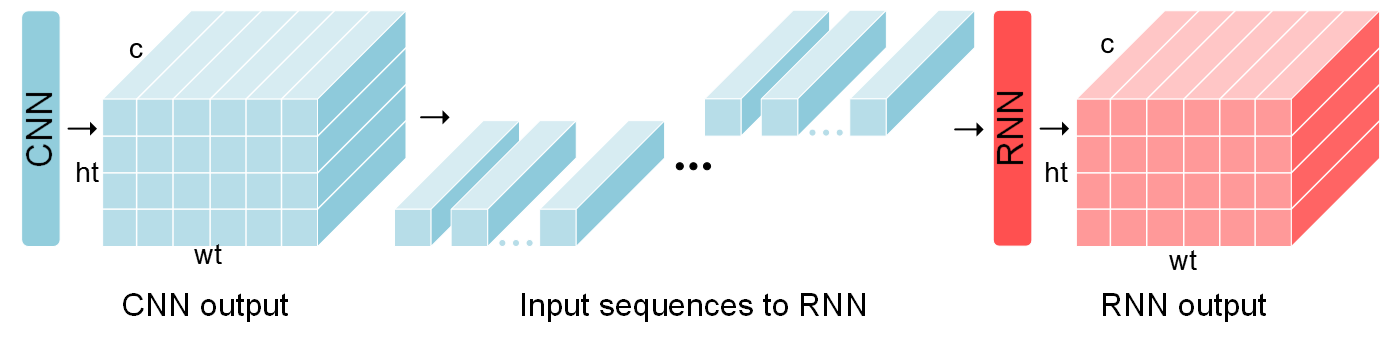}
\end{center}
\vspace{-5mm}
\caption{Inserting an RNN layer after a CNN layer.
The output feature map of the CNN layer is divided into cells and passed to the RNN layer in sequence, one cell at a time.
The RNN outputs are stacked to form a volumetric feature map with similar size as the CNN output feature map.}
\label{fig:rnn}
\end{figure}

Additionally, we add RNN layers at the last few feature maps as shown in~\figref{fig:rnn}.
Each CNN output feature map is converted into a sequence of cells along the width and height dimension.
The sequence length is $wt \times ht$; where $wt$ and $ht$ are the width and height of the feature map, respectively.
We implement the RNN layer using a unidirectional GRU with the hidden size equal to $c$. 
For each input cell, the RNN layer returns an output cell of the same size. These output cells are stacked along the width and height dimension to form a volumetric feature map with the same dimension as the output of the preceding CNN layer. 
In addition, since RV images are contiguous along the width dimension,  we circularly pad the CNN output along this dimension to better initialize the hidden states of the RNN layer.
The RNN layers help to learn spatial relationships of objects in the azimuth direction, which are strongly present in RV images.
For example, \emph{sidewalk} is often next to \emph{road} and \emph{terrain} is often next to \emph{vegetation}. 
\vspace{-1mm}
\mypar{Bird's-eye view network.} We implement the PolarNet~\cite{Polarnet} method, which is currently the state-of-the-art method for LiDAR semantic segmentation in BEV.
This network uses a U-Net architecture and outputs a voxel-level segmentation given polar-pillar features.

\noindent For both networks, we perform the following point data augmentation steps: scaling, flipping in the x and y directions, jitter, and rotation in the yaw direction.

\squeeze
\vspace{-1mm}
\subsection{Assertion-guided Point Sampling}
\squeeze
Once we obtain the initial class predictions from the aforementioned networks, we sample points for training and inference for the point head.
The sampling is based on an assertion-guided approach to capture uncertain  points where the multi-view networks disagree with each other on the class predictions.

Given a LiDAR scan with $N$ points, we define the point cloud to be $\mathbf{X}=\{\mathbf{x}_{i}\}|_{i=1}^{N}$, where $\mathbf{x} = (x, y, z, intensity)$. The objective is to obtain the final prediction label vector $\mathbf{y} \in \mathbb{R}^{K}$ for all points, where $K$ is the number of classes.

We define the normalized class predictions of RV and BEV networks, parameterized by $\theta_{rv}$ and $\theta_{bev}$, to be $f \in \mathbb{R}^{N \times K} : f(\mathbf{x} | \mathbf{X}, \theta_{rv})$ and $g \in \mathbb{R}^{N \times K}: g(\mathbf{x}|\mathbf{X}, \theta_{bev})$, respectively. 
Given a point $\mathbf{x}_{j}$, the uncertainty $u_{j} \in \{0, 1\}$  is defined as:
\begin{equation}
u_{i} =    \begin{cases}
     1 & \text{if} \quad s(f_i, g_i) <= \tau \\
     0 & \text{otherwise}
   \end{cases}
   \label{eq:uncertainty}
\end{equation}
where $s(\cdot, \cdot)$ is the cosine similarity score  and $\tau$ is a design parameter for the uncertainty threshold.

Aside from multi-view disagreements, our framework can be easily extended to any type of assertion such as class entropy, high network loss, etc. 
While there exists literature about model assertions\cite{ModelAssertion}, we are novel in which we apply the checks during training and inference of an existing dataset while theirs does it for Active Learning.

\squeeze
\vspace{-1mm}
\subsection{Point Head Architecture} \label{subsec:point_head_arch}
\begin{figure}
\begin{center}
\includegraphics[width = 0.5\textwidth]{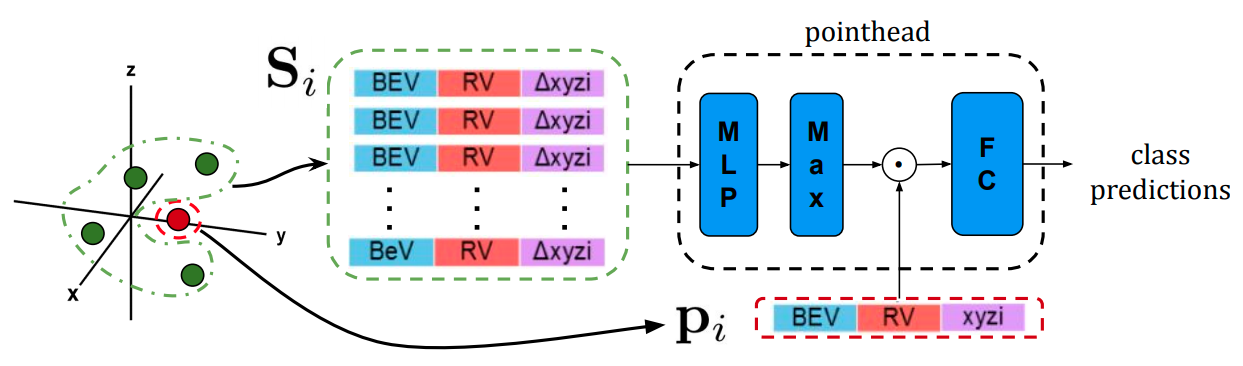}
\end{center}
\vspace{-6mm}
\caption{Point head architecture which obtains class predictions given point-level features, $\mathbf{p}_{i}$, and its neighborhood set features, $\mathbf{S}_{i}$}
\label{fig:pointhead}
\end{figure}

\squeeze
Given an uncertain point, we extract relevant features to feed to the point head for the final label prediction. 
We define its point-level features as the concatenation of the individual network's normalized class scores and  the raw point data\footnote{Note that these projection-specific features can be intermediate features from the networks' feature maps, but for simplicity, we use normalized class scores.}:
\begin{equation}
\mathbf{p}_{i} = [f_{i} \quad g_{i} \quad \mathbf{x}_{i}]
\label{eq:point_features}
\end{equation} 
We also select the point-level features of the neighbouring points for additional context.
Given $\mathcal{N}(\mathbf{x}_{i})$ to be the set of $n$ neighboring points for the $i$th point, the neighborhood set features are defined as:
\begin{eqnarray}
\mathbf{S}_{i} =
\begin{bmatrix}
f_{k_1} & g_{k_1} & \phi(\mathbf{x}_{i}, \mathbf{x}_{k_1}) \\
\vdots & \vdots & \vdots \\
f_{k_n} & g_{k_n} & \phi(\mathbf{x}_{i}, \mathbf{x}_{k_n})
\end{bmatrix}
\label{eq:neighbor}
\end{eqnarray}
where the point feature $\phi$ is the relative distance between the neighboring point $k$ and the uncertain point $i$.
We perform k-Nearest Neighbor (using KDTree) on the raw points to extract the neighboring points, with $n$ as a hyperparameter.
We pass these point features and neighborhood set features to a point head to obtain the final predictions.
The point head class predictions $\mathbf{h}_{i} \in \mathbb{R}^{K}$ for the point $\mathbf{x}_i$ are as follows:
\begin{equation}
\mathbf{h}_{i} = \text{pointhead}(\mathbf{p}_{i}, \mathbf{S}_{i})
\label{eq:pointhead}
\end{equation}
The point-head architecture is shown in \figref{fig:pointhead}.
It comprises of an MLP, a max pooling layer, and a fully connected (FC) layer.
$\mathbf{S}_{i}$ is processed independently through an MLP and maxpooled to obtain a new point feature which is then concatenated with $\mathbf{p}_{i}$. 
This is similar to the local point embedder proposed in~\cite{LocalPointEmbedder} except that they use this architecture at an earlier stage of their pipeline to extract features from raw point clouds. 
Instead, we try to learn to predict the the final class labels from already meaningful point features from the RV and BEV networks.
We use the CE loss function to train our point-head network.

During training, we randomly select a batch of uncertain points to train the point head at every iteration.
During inference, we select all the points that fall under the threshold for a second round of predictions using our trained point head.
The final prediction label vector is as follows:
\begin{equation}
\mathbf{y}_{i} =    \begin{cases}
     \mathbf{h}_{i} & u_{i} = 1 \\
     \sqrt{f_{i} \cdot g_{i}} & u_{i} = 0
   \end{cases}
   \label{eq:final_prediction}
\end{equation}
For a point whose uncertainty is above the threshold, we make a class prediction using the geometric mean of the predictions from the RV and BEV networks.  

Our point head consists of 18.5K parameters with its MLP having only 3 hidden layers, making it light-weight.  
Given 10K points, which is a typical number of uncertain points in SemanticKITTI, it would only have a total of 1 GFLOPs.
This is roughly $0.2\%$ of the RV network's GFLOPs for a single LiDAR scan and can be considered negligible if we consider the whole AMVNet pipeline.

%% file: sections/experiments.tex
% !TEX root = ../amvnet.tex
\begin{table*} \centering
\setlength\tabcolsep{4pt} % Decrease this to reduce margins in each cell.
\footnotesize
    \begin{tabularx}{1.01\textwidth}{L{2cm}ccccccccccccccccccc|cc}
        Method & \rot{car} & \rot{bicycle} & \rot{motorcycle} & \rot{truck} & \rot{other-vehicle} & \rot{person} & \rot{bicyclist} & \rot{motorcyclist} & \rot{road} & \rot{parking} & \rot{sidewalk} & \rot{other-ground} & \rot{building} & \rot{fence} & \rot{vegetation} &  \rot{trunk} & \rot{terrain} & \rot{pole} & \rot{traffic-sign} & \rot{\textbf{mIOU}} \\
        \midrule
         PointNet~\cite{pointnet}             & 46.3 & 1.3  & 0.3  &  0.1 & 0.8  & 0.2  & 0.2 & 0.0  &   61.6  & 15.8 & 35.7 & 1.4 & 41.4 & 12.9 & 31.0& 4.6& 17.6& 2.4 & 3.7 & 14.6 \\
         RandLANet~\cite{RandLA-Net}             &94.0& 19.8  & 21.4  & 42.7 & 38.7  & 47.5   & 48.8  & 4.6  & 90.4  & 56.9 & 67.9 & 15.5 & 81.1 & 49.7 & 78.3 & 60.3 & 59.0 & 44.2 & 38.1 & 50.3\\\hline
         RangeNet++~\cite{Rangenet++}             & 91.4 &  25.7 & 34.4  & 25.7  & 23.0  &38.3   &38.8  & 4.8  & \textbf{91.8}  & 65.0 & 75.2 & 27.8 & 87.4& 58.6 & 80.5 & 55.1 & 64.6 & 47.9 & 55.9 &52.2 \\
         SalsaNext~\cite{Salsanext}             & 91.9 & 48.3  & 38.6  & 38.9  & 31.9  & 60.2  & 59.0  & 19.4   & 91.7  & 63.7 & 75.8 & 29.1& 90.2& 64.2& 81.8& 63.6& 66.5& 54.3& 62.1& 59.5\\
KPRNet~\cite{Kprnet}  &  95.5& 54.1  & 47.9  & 23.6  & 42.6  & 65.9  & 65.0 & 16.5  & \textbf{93.2}  & \textbf{73.9} & \textbf{80.6} & 30.2 & 91.7 & 68.4 & \textbf{85.7} & 69.8 & \textbf{71.2} & 58.7 & 64.1 & 63.1\\\hline
                  PolarNet~\cite{Polarnet}             & 93.8 & 40.3  & 30.1  & 22.9  & 28.5  & 43.2  & 40.2 &  5.6 & 90.8  & 61.7 & 74.4 & 21.7 & 90.0 & 61.3 & 84.0 & 65.5 & 67.8 & 51.8  & 57.5 & 54.3 \\ \hline
         Cylinder3D~
         \cite{Cylinder3d} & 96.1  & 54.2  & 47.6  & 38.6  & 45.0  & 65.1  & 63.5 & 13.6  & 91.2  & 62.2& 75.2& 18.7& 89.6& 61.6& 85.4& 69.7& 69.3& 62.6& 64.7& 61.8\\
          FusionNet~\cite{FusionNet} & 95.3 & 47.5  & 37.7  & 41.8  & 34.5  & 59.5   & 56.8  & 11.9  & \textbf{91.8}  & 68.8 & \textbf{77.1}& \textbf{30.8}& \textbf{92.5} & \textbf{69.4} & 84.5 & 69.8 & 68.5& 60.4& 66.5& 61.3 \\
         SPVNAS~\cite{SPVNAS} & \textbf{97.2} & 50.6  & \textbf{50.4}  &\textbf{56.6}   & \textbf{58.0}  & \textbf{67.4}   & \textbf{67.1} & \textbf{50.3}  &90.2   & 67.6  & 75.4 & 21.8 & 91.6 & 66.9 & \textbf{86.1} & \textbf{73.4} & 71.0 & \textbf{64.3} & \textbf{67.3} & \textbf{67.0} \\ \hline
         TornadoNet~\cite{Tornadonet} & 94.2 & \textbf{55.7}  & 48.1  & 40.0  & 38.2  & 63.6  & 60.1 & \textbf{34.9}  &89.7   &66.3 &74.5 & 28.7& 91.3 & 65.6& 85.6& 67.0& \textbf{71.5} & 58.0& 65.9& 63.1 &\\
        \textbf{AMVNet}             & \textbf{96.2} & \textbf{59.9}  & \textbf{54.2} & \textbf{48.8}  & \textbf{45.7} & \textbf{71.0}  &  \textbf{65.7} & 11.0  & 90.1 & \textbf{71.0}  & 75.8 & \textbf{32.4} & \textbf{92.4} &\textbf{ 69.1} & 85.6 & \textbf{71.7} & 69.6 & \textbf{62.7}& \textbf{67.2}& \textbf{65.3}   \\
        \bottomrule
    \end{tabularx}
    \vspace{-3mm}
    \caption{Class-wise IOU on the test set of the SemanticKITTI official leaderboard. The methods are grouped, from top to bottom, as point-based networks, RV networks, BEV networks, 3D voxel partition-based networks and multi-view fusion networks. Best two values for each class are in \textbf{bold}. All IOU scores are given in percentage ($\%$).
    }
\label{table:semkitti_test}
\end{table*}
\begin{table*} \centering
\setlength\tabcolsep{5pt}
\footnotesize
    \begin{tabular}{lcccccccccccccccc|cc}
        Method & \rot{barrier} & \rot{bicycle} & \rot{bus} & \rot{car} & \rot{constr.veh} & \rot{motorcycle} & \rot{pedestrian} & \rot{traffic cone} & \rot{trailer} & \rot{truck} & \rot{driv.surf.} & \rot{other flat} & \rot{sidewalk} & \rot{terrain} & \rot{manmade} &  \rot{vegetation} & \rot{\textbf{mIOU}} & \rot{\textbf{FW IOU}$^{\ddagger}$}\\
        \midrule
         PolarNet~\cite{Polarnet}             &  72.2 & 16.8  & 77.0   & \textbf{86.5}  & 51.1  & 69.7  & 64.8 &   54.1 &  69.7 & 63.4 & 96.6  & \textbf{67.1} & 77.7 & 72.1& 87.1& 84.4& 69.4 &  87.3\\
         RV Net$^{\dag}$             & 77.9 & 30.2  & 81.9  & 83.2  & 60.9  & 78.9  & 69.0 &  65.8  & 81.5  & 64.1 & 97.0 & 66.2 & 77.6 & 73.8 & 88.2 & 86.2 & 73.9 & 88.2\\
         \textbf{AMVNet}             & \textbf{79.8} & \textbf{32.4}  & \textbf{82.2}  & 86.4  & \textbf{62.5}  & \textbf{81.9}  & \textbf{75.3} & \textbf{72.3}  & \textbf{83.5}  & \textbf{65.1} & \textbf{97.4} & 67.0  & \textbf{78.8} & \textbf{74.6} & \textbf{90.8} & \textbf{87.9} & \textbf{76.1} & \textbf{89.5}\\
        \bottomrule
    \end{tabular}
    \vspace{-3mm}
    \caption{Class-wise IOU on the test set of the nuScenes LiDAR semantic segmentation challenge.
    $^{\dag}$An implementation of an RV network as described in~\secref{sec:main_networks} without the RNN. $^{\ddagger}$FW IOU denotes Frequency Weighted IOU. Best values for each class are in \textbf{bold}.
    }
\label{table:nuscenes_test}
\end{table*}
\squeeze
\section{Experiments}
\label{sec:experiments}
\squeeze
To demonstrate the effectiveness of our proposed methods, we present benchmark evaluations and in-depth analyses through ablation studies and visualizations.
\squeeze
\subsection{Datasets}
\squeeze
We evaluate our method on two benchmark datasets SemanticKITTI~\cite{Semantickitti} and nuScenes~\cite{Nuscenes}. 

\mypar{SemanticKITTI.}
This dataset is based on the odometry dataset of the KITTI Vision Benchmark \cite{KITTI}, where it consists of 43551 LiDAR scans from 22 sequences collected in a city in Germany; in which 10 scenes (19130 scans) are used for training, 1 scene (4071 scans) for validation, and 10 scenes (20351 scans) for testing.
The dataset has been collected using a Velodyne HDL-64E sensor with horizontal angular resolution of 0.08 to 0.35, and has 64 beams vertically.
SemanticKITTI provides up to 28 classes, but a high-level label set of 19 classes is used for the official evaluation using a single scan.
Each scan has approximately 130K points. 
This dataset presents challenges on rare classes such as \emph{motorcyclists} and \emph{other-ground} due to the limited training examples.
Furthermore, classes like \emph{pedestrian}, \emph{bicyclist} and \emph{motorcyclists} are difficult to differentiate semantically.

\mypar{nuScenes.} This dataset is a multi-modal dataset for 3D object detection and tracking. Recently, nuScenes released point-level annotations for its 1000 scenes\footnote{https://www.nuscenes.org/nuscenes\#lidarseg}.
Each scene is 20s long, collected from different areas of Boston and Singapore.
It has 28130 samples for training, 6019 for validation and  6008 for testing.
The dataset uses a Velodyne HDL-32E sensor with horizontal angular resolution of 0.1 to 0.4, and has 32 beams vertically.
The beam information for each point is also provided in the dataset.
The annotated dataset provides up to 32 classes, but similar to SemanticKITTI, a high-level label set of 16 classes is used for the official evaluation.
Each scan typically has around 34K points.
Like SemanticKITTI, this dataset poses challenges based on class imbalance. In particular,  classes like \emph{bicycles} and \emph{construction vehicles} have relatively limited training data.
Moreover, nuScenes is challenging as it encompasses different locations and diverse weather conditions with overall more objects.
nuScenes is also less dense as the sensor has fewer number of beams and lower horizontal angular resolution.
\squeeze
\subsection{Implementation Details}
\label{sec:implementation_details}
\squeeze
In this subsection, we provide dataset-specific details of our networks. 
\mypar{SemanticKITTI.} Our RV network accepts a $64 \times 4096$ spherical projection  and uses feature maps from strides 4 to 128 to pass through as inputs to the decoder network.
The RNN layers are injected in strides 64 and 128.
The RV network is trained for 160 epochs with a batch size of 32.
Our BEV network uses a polar grid size of $480 \times 360 \times 32$  to cover a grid space of radius $50m$  and height between $(-3m, 1.5m)$ where the height is relative to the LiDAR sensor.
It is trained for 40 epochs with a batch size of 8.
\vspace{-0.6mm}
\mypar{nuScenes.} Our RV network uses a $32 \times 1920$ cylindrical input and uses strides 4 to 64 for the segmentation head.
It is trained for 400 epochs with a batch size of 64.
Our BEV network uses a polar grid size of $480 \times 360 \times 32$ to cover a grid space of radius $50m$ and height between $(-2m, 4m)$.
It is also trained for 40 epochs with a batch size of 8.

\noindent For both datasets, the RV and BEV networks are trained using one-cycle learning rate schedule and AdamW optimizer.
We also include class-specific weighting in the loss functions which are proportional to the square-root inverse of class frequencies to handle the class imbalance.
For point head training, we use 20 epochs with a batch size of 256 points, an uncertainty threshold of $\tau=0.85$ and $n=15$ point neighbors. The point head is trained using one-cycle learning rate schedule and SGD optimizer.
Our networks are trained independently, but can be explored to be end-to-end in future work.

\begin{table*} \centering
\setlength\tabcolsep{4pt} % Decrease this to reduce margins in each cell.
\footnotesize
    \begin{tabularx}{1.05\textwidth}{L{1.45cm}ccccccccccccccccccc|cc}
        Method & \rot{car} & \rot{bicycle} & \rot{motorcycle} & \rot{truck} & \rot{other-vehicle} & \rot{person} & \rot{bicyclist} & \rot{motorcyclist} & \rot{road} & \rot{parking} & \rot{sidewalk} & \rot{other-ground} & \rot{building} & \rot{fence} & \rot{vegetation} &  \rot{trunk} & \rot{terrain} & \rot{pole} & \rot{traffic-sign} & \rot{\textbf{mIOU}} & \rot{\textbf{FW IOU}}\\
        \midrule
         BEV & 94.0 & 30.0  & 56.0  & 64.6  & 42.9 & 60.6 & 77.3 & 0.0 & 93.5 & 41.2 & 79.3 & 0.19 & 88.6 & 46.3 & 86.5 & 56.0 & 73.2 & 62.5 & 45.8 & 58.9 & 83.6 \\
         RV  & 90.4  & 31.9  & 57.6  & 79.8 & 45.7 & 61.9 & 64.9& 0.0 & 95.3 & 48.9 & 81.8 & \textbf{0.79} & 85.3 & 59.7 & 84.1&  58.8& 69.9& 53.4& 44.7& 59.6 & 82.9 \\
        \textbf{AMVNet}   & \textbf{95.6}  & \textbf{48.8}   & \textbf{65.4}   & \textbf{88.7}  & \textbf{54.8} & \textbf{70.8} & \textbf{86.2} & 0.0 & \textbf{95.5} & \textbf{53.9} & \textbf{83.2} & 0.15 & \textbf{90.9} & \textbf{62.1} & \textbf{87.9} & \textbf{66.8} & \textbf{74.2} & \textbf{64.7} & \textbf{49.3} & \textbf{65.2} & \textbf{86.0} \\
        \bottomrule
    \end{tabularx}
    \vspace{-3mm}
    \caption{Class-wise IOU on the validation set of the SemanticKITTI between our individual networks and AMVNET.} 
\label{table:semkitti_val}
\end{table*}

\squeeze
\subsection{Experimental results}
\label{sec:test_results}
\squeeze
Here we present AMVNet results on benchmark datasets submitted to their respective evaluation servers.
\vspace{-.3mm}
\mypar{SemanticKITTI.} \tabref{table:semkitti_test} shows the test results of our proposed method compared to various LiDAR segmentation approaches in the recent years.
As seen in the table, we have achieved the second highest mIOU score just behind SPVNAS~\cite{SPVNAS} and is +2.2\% better than the current best multi-view approach~\cite{Tornadonet}.
We also performed best in certain classes such as \emph{bicycle}, \emph{motorcycle}, \emph{person} and~\emph{other-ground}.

While SPVNAS provides very good results, we want to point out that their key contributions are a 3D neural architecture search (3D-NAS) and a sparse point-voxel convolutions (SPVConv). Differently, our scope is presenting a better option for leveraging the strengths of multiple views instead of concentrating on network optimization. 
Looking into the class-wise IOU, our method performs better than SPVNAS in 8/19 categories and is comparable in 6/19.
Nevertheless, both approaches are complementary and can be applied together to achieve better performance.

\mypar{nuScenes.} The dataset is relatively new so we ran our re-implementation of the PolarNet code and a variant of our range view network as baselines to compare with our AMVNet as shown in \tabref{table:nuscenes_test}. 
AMVNet performs better compared to the best baseline by $+3\%$ mIOU.
AMVNet also outperforms the baselines in most classes, with significantly better performance in \emph{pedestrian} (+6), \emph{traffic cone} (+6), and \emph{motorcycle} (+3).
Our AMVNet also achieved a higher mIOU in nuScenes compared to SemanticKITTI.
One reason may be the fewer coarse-grained classes, with no distinction between a \emph{person}, \emph{bicyclist} and \emph{motorcyclist} as well a \emph{fence}, \emph{pole} and \emph{traffic sign}, where network confusions are likely to occur.

\squeeze
\subsection{Further Analysis}
\label{sec:ablation_studies}
\squeeze
In this subsection, we run more analyses on different aspects of our method.

\begin{figure}
\centering
\begin{subfigure}[b]{.23\textwidth}
  \centering
  % include first image
  \includegraphics[width=\linewidth]{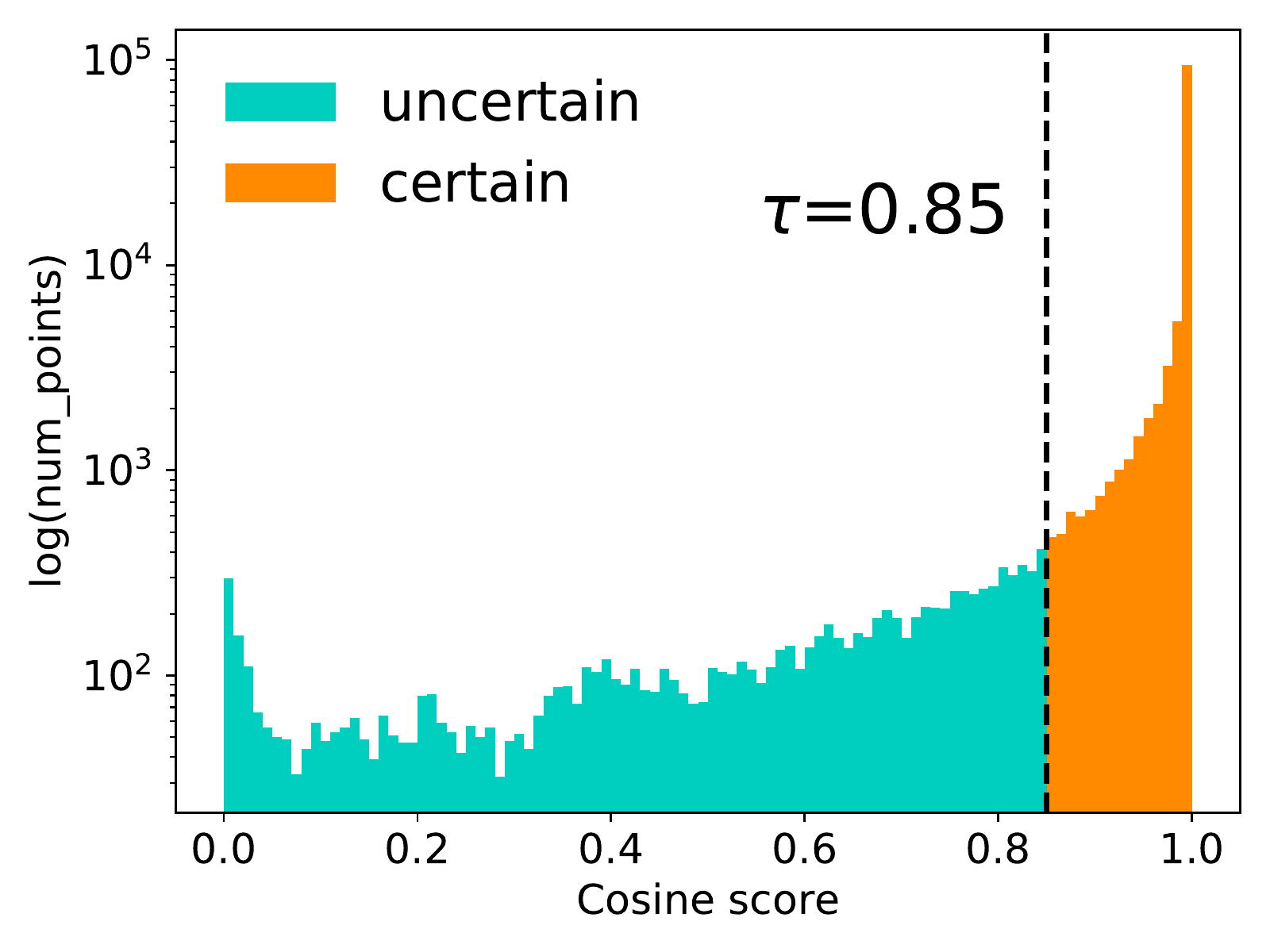}
  \vspace{-6mm}
  \caption{SemanticKITTI}
  \label{fig:sub-first}
\end{subfigure}
\begin{subfigure}[b]{.23\textwidth}
  \centering
  % include second image
  \includegraphics[width=\linewidth]{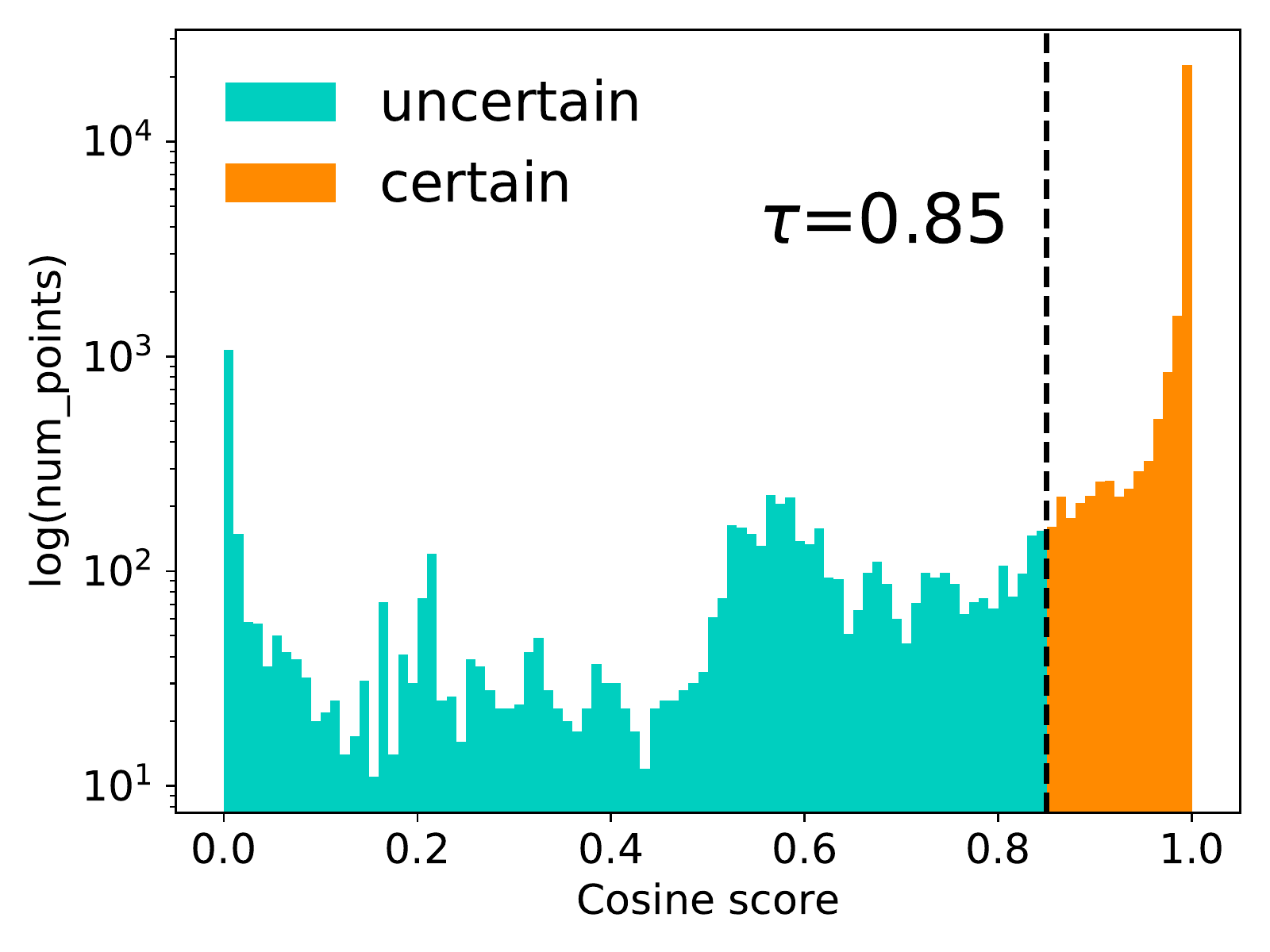}
  \vspace{-6mm}
  \caption{nuScenes}
  \label{fig:sub-second}
\end{subfigure}
\vspace{-3mm}
\caption{Histogram plots of cosine similarity between RV and BEV label predictions of all points (log-scaled) in one LiDAR scan.
Points below the threshold, which are flagged as uncertain, are only $10$ to $20\%$ of all points.}
\label{fig:fig_histogram}
\end{figure}

\mypar{RV and BEV disagreements.} We examine closely the cosine similarity of RV and BEV predictions given a LiDAR scan in~\figref{fig:fig_histogram}.
We see that in most points, RV and BEV networks already agree in the label predictions.
On average, only $\approx$10\% in SemanticKITTI ($\approx20\%$ in nuScenes) of total number of points fall below  $\tau=0.85$, classified as uncertain,  and are fed to the point head.
This largely reduces the point-level computations during inference.

\mypar{Ensemble vs. AMVNet.}  \tabref{table:ensemble_comparison} compares our approach on the individual network's performance and combining the predictions of both networks in a simple ensemble.
For our ensemble, we explore different ways (e.g. arithmetic mean, geometric mean and max) to combine the class predictions of both networks and achieved best results using the geometric mean.
As shown, combining both predictions through an ensemble provides better performance than the individual networks which clearly suggests that both networks complement each other.
More importantly, our proposed approach is better than just the ensemble, with a difference of approximately 2.5 mIOU for SemanticKITTI. 
\begin{table} \centering
\footnotesize
    \begin{tabular}{lcccc}
        Dataset & RV & BEV & Ensemble & AMVNet \\
        \midrule
         SemanticKitti         & 59.6 & 58.9  & 62.6  & \textbf{65.2}  \\
         nuScenes              & 74.2 & 69.5  & 76.0  & \textbf{77.2}   \\
        \bottomrule
    \end{tabular}
    \vspace{-3mm}
    \caption{mIOU comparison between individual networks, ensemble and AMVNet in the validation set.}
\label{table:ensemble_comparison}
\end{table}

\begin{figure}
\centering
\begin{subfigure}[b]{.23\textwidth}
  \centering
  % include first image
  \includegraphics[width=\linewidth]{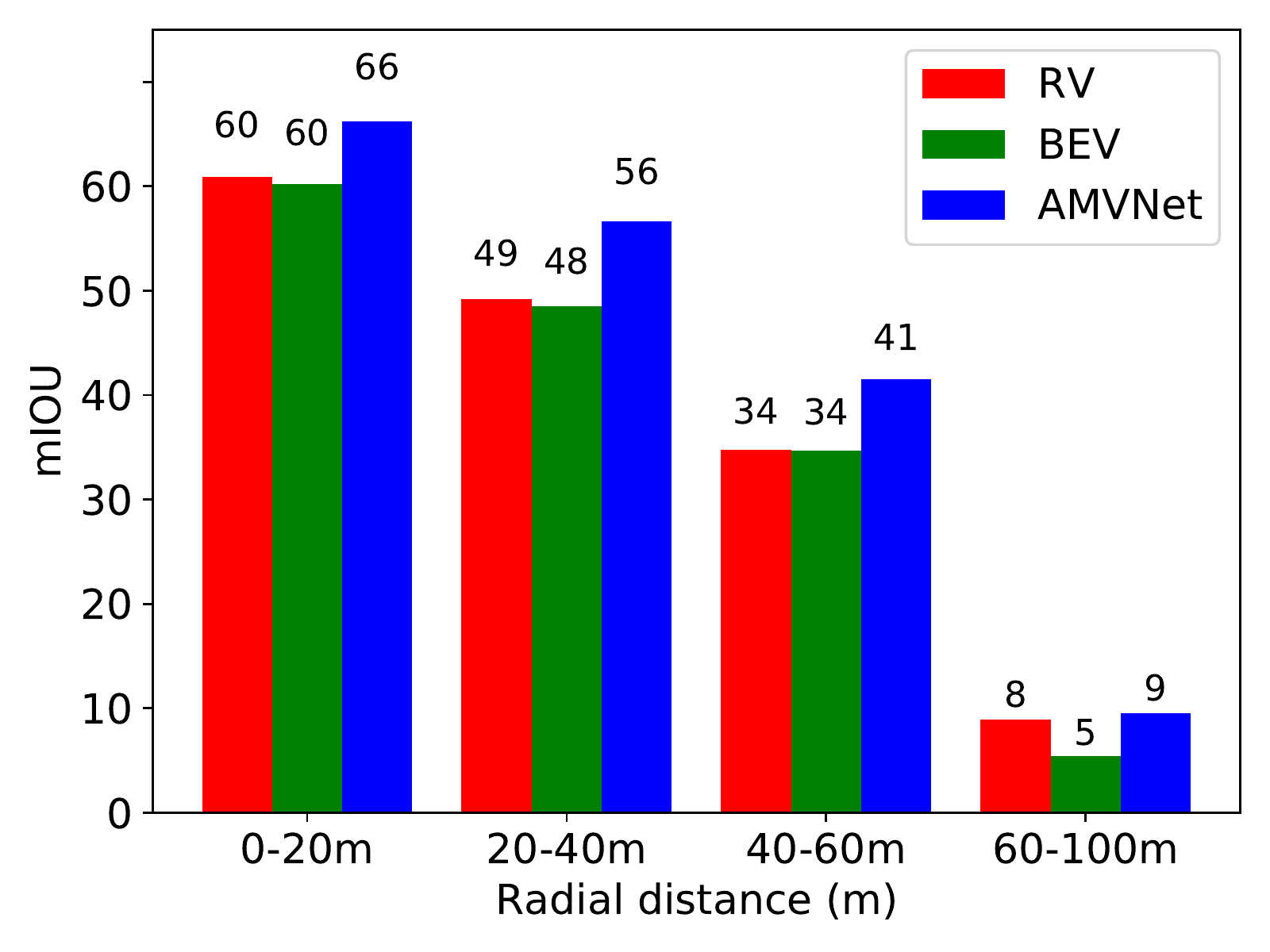}
    \vspace{-6mm}
  \caption{SemanticKITTI}
  \label{fig:sub-first}
\end{subfigure}
\begin{subfigure}[b]{.23\textwidth}
  \centering
  % include second image
  \includegraphics[width=\linewidth]{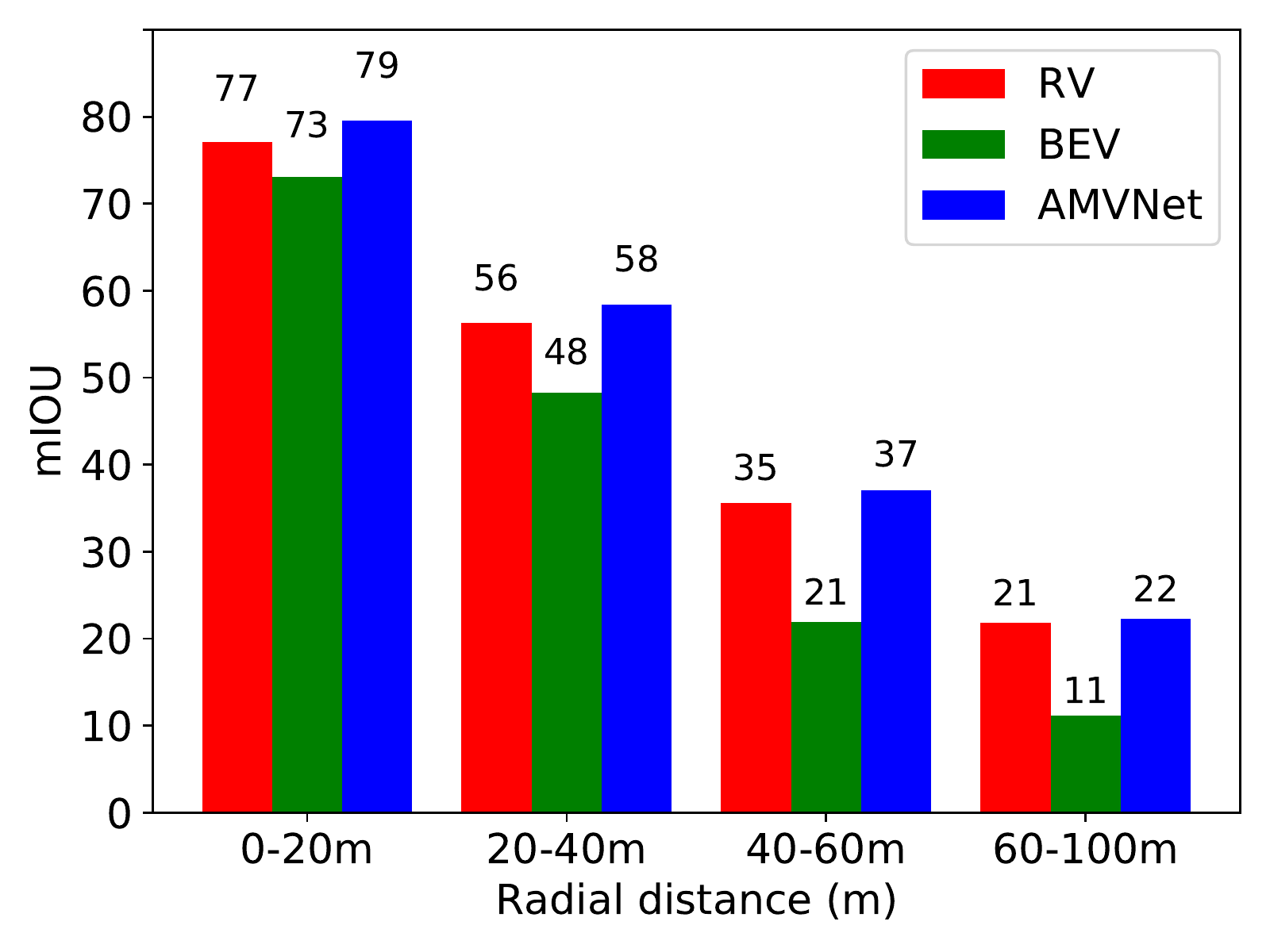}
    \vspace{-6mm}
  \caption{nuScenes}
  \label{fig:sub-second}
\end{subfigure}
\vspace{-3mm}
\caption{mIOU at varying radial distance stratifications for the validation sets.}
\label{fig:strata}
\end{figure}

\begin{figure*}
\centering
\begin{subfigure}[b]{.47\textwidth}
  \centering
  % include first image
  \includegraphics[width=\linewidth]{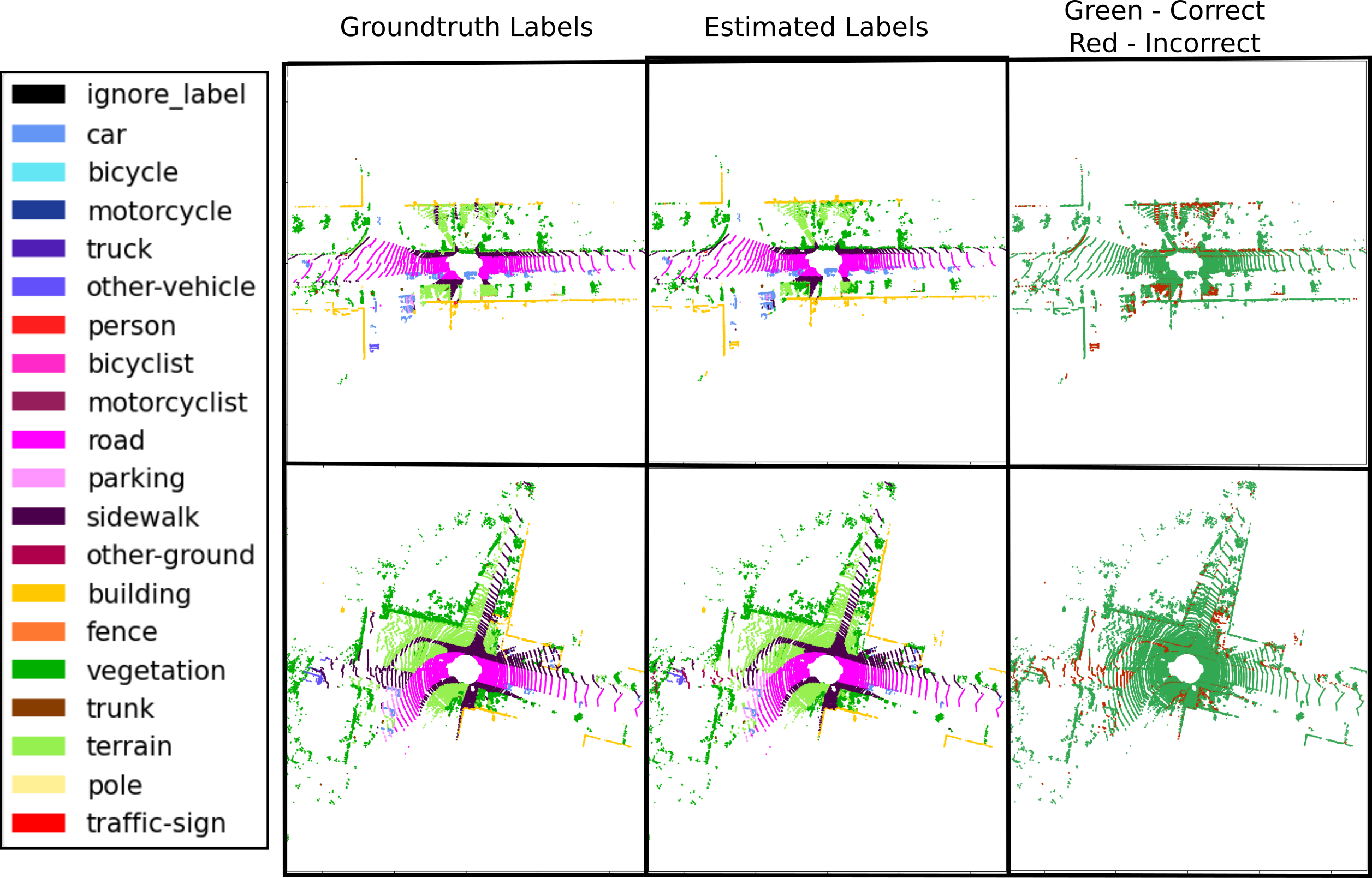}
  \caption{SemanticKITTI}
  \label{fig:sub-first}
\end{subfigure}
\begin{subfigure}[b]{.47\textwidth}
  \centering
  % include second image
  \includegraphics[width=\linewidth]{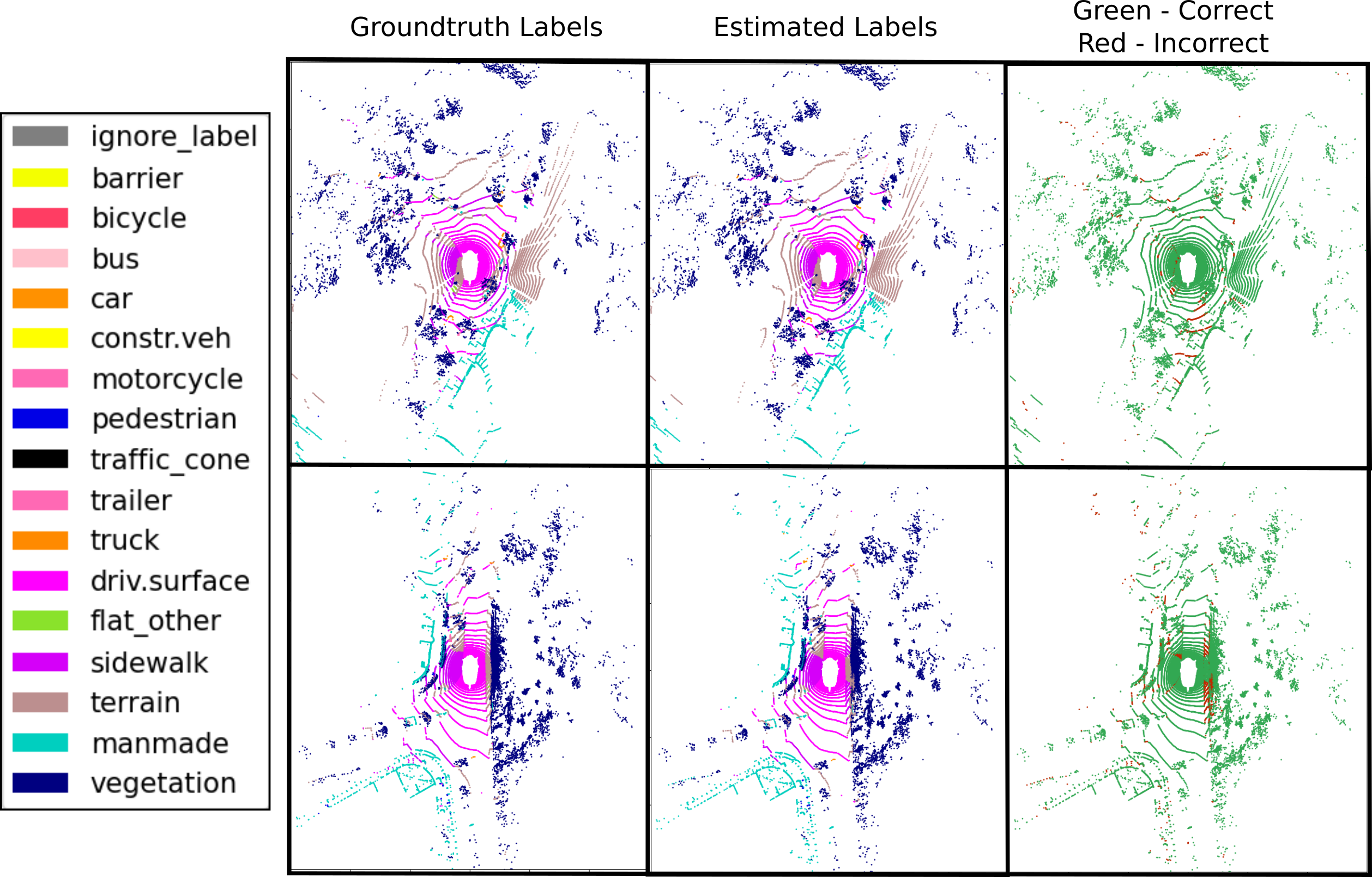}
  \caption{nuScenes}
  \label{fig:sub-second}
\end{subfigure}
\vspace{-3mm}
\caption{AMVNet point cloud renderings in top view given sample LiDAR scans taken from benchmarks' validation set.
}
\label{fig:visuals}
\end{figure*}

\mypar{Radial distance stratification.} We check the mIOU at varying stratifications based on the radial distance from the ego-vehicle. 
\figref{fig:strata} shows the comparisons.
The overall mIOU of our AMVNet is better compared to the individual networks in all stratas in SemanticKITTI.
This suggests that in all stratas the individual networks complement each other.
For nuScenes, the BEV model is generally weaker compared to RV in all stratas.
This leads to smaller improvements in the AMVNet (1-$2\%$ mIOU) compared to the individual networks.

\mypar{Class-wise analysis.}\tabref{table:semkitti_val} shows a more in-depth class-wise comparison between the individual networks used and AMVNet.
It shows that the BEV model is better in certain classes such as \emph{car}, \emph{vegetation}, \emph{terrain}, \emph{pole} and \emph{traffic sign} while the RV model is better elsewhere.
AMVNet is able to leverage the strengths of the individual networks and obtains best IOU across all classes except for \emph{other-ground}.
Note that the \emph{other-ground} has very limited number of points in the validation split, hence not conclusive on the overall performance. 

\begin{figure}
\centering
\begin{subfigure}[b]{.23\textwidth}
  \centering
  % include first image
  \includegraphics[width=\linewidth]{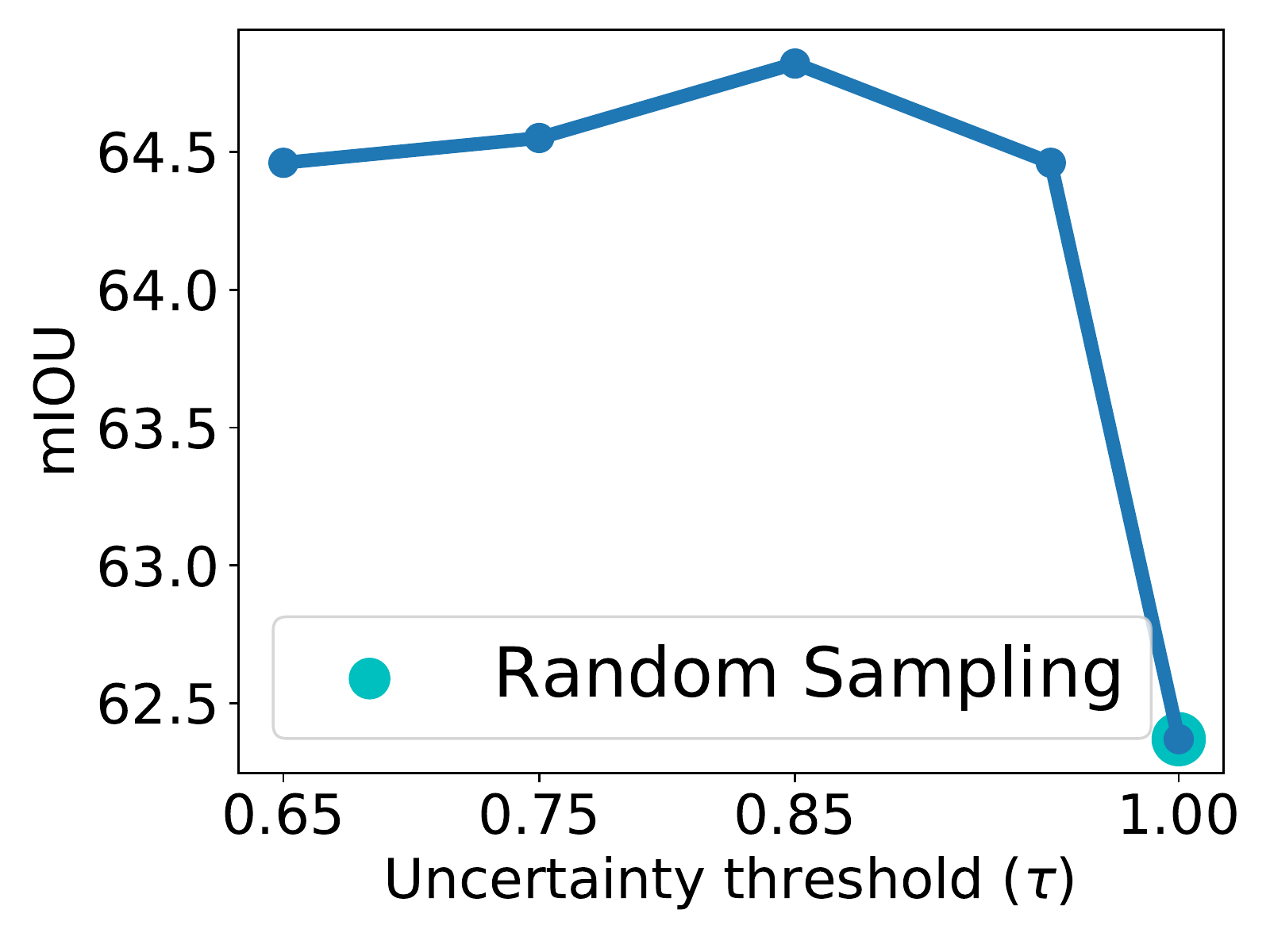}
    \vspace{-6mm}
  \caption{$\tau$}
  \label{fig:tau}
\end{subfigure}
\begin{subfigure}[b]{.23\textwidth}
  \centering
  % include second image
  \includegraphics[width=\linewidth]{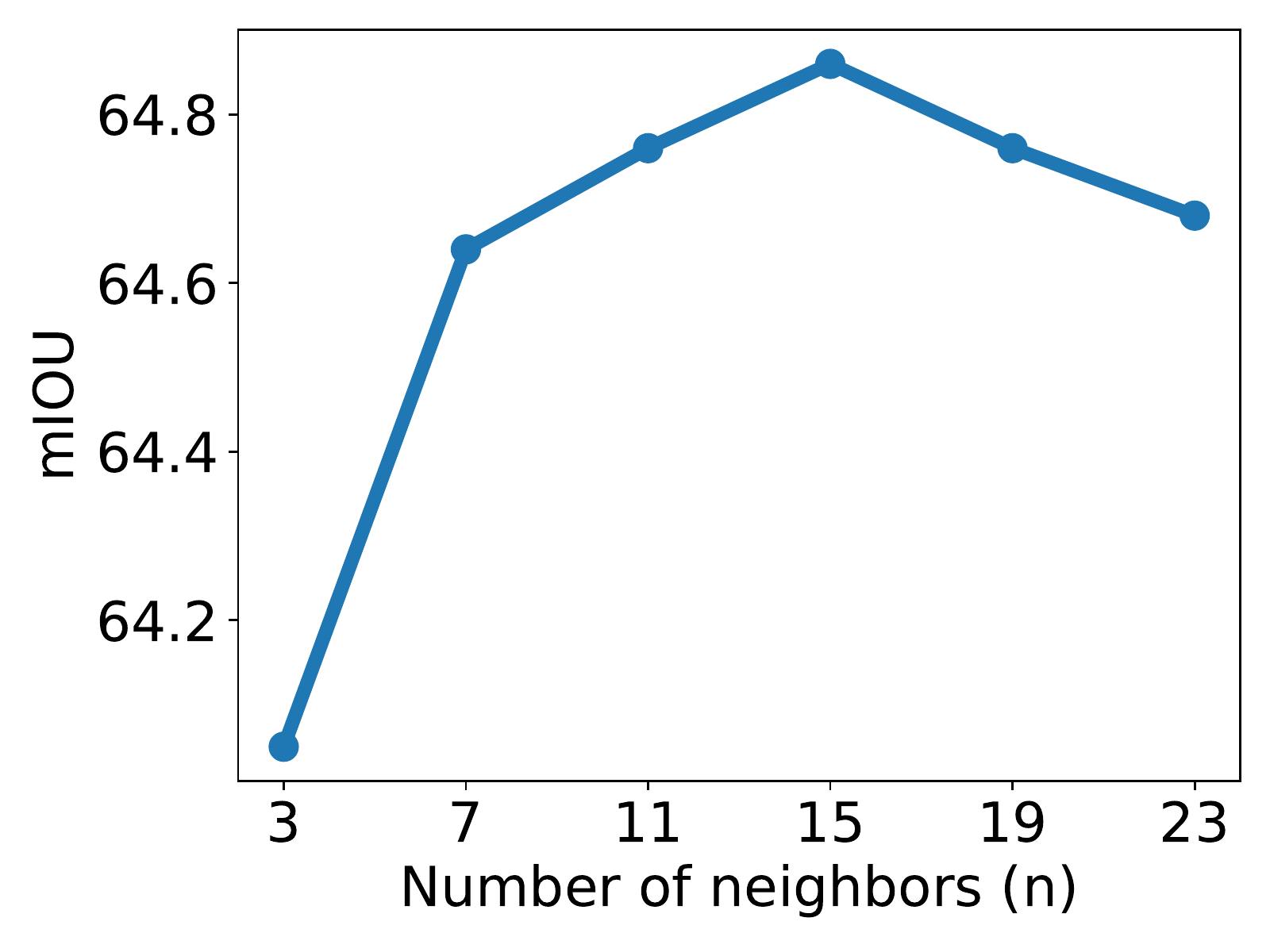}
    \vspace{-6mm}
  \caption{neighborhood points}
  \label{fig:neighborhood}
\end{subfigure}
\vspace{-3mm}
\caption{Hyperparameter sweep in the SemanticKITTI validation set.}
\label{fig:fig_parametersweep}
\end{figure}

\mypar{Varying threshold for point sampling.} We investigate the effect of varying the thresholds in determining uncertainty of a point as shown in~\figref{fig:tau}.
At $\tau=1$, which means that we are randomly sampling points for training, we see a relatively low mIOU. This suggests that training with random points is just as good as using an ensemble of the two networks.
This strengthens the need of our assertion-guided sampling.

\mypar{Varying number of neighbors.} We also investigate the effect of varying the number of neighbors $n$, for the set features $\mathbf{S}$, in Eq.~\eqref{eq:neighbor} which we feed to the point head.
\figref{fig:neighborhood} shows the mIOU results, which shows that adding neighborhood information helps in the mIOU performance with a huge gap between $n=3$ and $n=7$.
$n=15$ gives the best results.

\mypar{RV enhancements.} We investigate the impact of the specific enhancements made in our RV network.
\tabref{table:ablation_studies} shows the contribution of both adding RNN layers and using class-weighting for losses. We see that in both enhancements, we obtained improved results by $1\%$ mIOU each. 
Combining them both increases the overall performance by $2\%$ mIOU points from the baseline.

\begin{table}
\centering
\footnotesize
    \begin{tabularx}{0.4\textwidth}{cccc}
         Range view & +RNN & +Class-Weighting &  mIOU \\
        \midrule
            \cmark     &  &   &  59.6   \\
            \cmark     &  \cmark &   &    60.7  \\
            \cmark     &   & \cmark  & 60.8     \\
            \cmark     &  \cmark &  \cmark & \textbf{61.7}  \\
        \bottomrule
    \end{tabularx}
    \caption{Ablation study on extensions for our RV network on the SemanticKITTI validation set.
    }
\label{table:ablation_studies}
\end{table}

\mypar{Qualitative Analysis.} \figref{fig:visuals} shows selected point cloud renderings of both datasets in the top view for our AMVNet compared to the groundtruth.
Overall, the output predictions of our AMVNet generally resemble the groundtruth. There are still some confusions that suggest further room for improvement. 
Particularly for SemanticKITTI, confusions are observed in  \emph{sidewalk} and \emph{parking}, \emph{terrain} and \emph{vegetation} and edges between \emph{sidewalk}/\emph{building}.
For nuScenes, confusions are also found in \emph{terrain} and \emph{vegetation}; and edges between \emph{sidewalk} and \emph{drivable surface}.

%% file: sections/conclusion.tex
% !TEX root = ../amvf.tex

\vspace{-2mm}
\squeeze
\section{Conclusion}
\label{sec:conclusion}
\squeeze
\vspace{-2mm}
In this paper we present a novel multi-view fusion method for LiDAR semantic segmentation for AVs, which leverages the strengths of RV and BEV methods. 
In this late fusion approach, we select uncertain points through an assertion-guided sampling strategy and extract relevant point features to feed the point head for a more robust prediction.  
By doing so, we are able to learn from difficult points and also provide some flexibility of having two decoupled networks in the system with optional processing using the point head. 
AMVNet achieved top results in two benchmark datasets.
Our work is easily extendable by adding more assertions, improving the point head to a more complex architecture (e.g. transformer~\cite{Transformer}), using ensembles of point heads, and improving the multi-view networks independently.

\vspace{+1mm}
\mypar{Acknowledgements.}
We thank Sourabh Vora for help in building the RV backbone architecture.
Oscar Beijbom and Alex H. Lang for support and guidance. 
Holger Caesar, Yiluan Guo and Kok Seang Tan for insightful review and discussion. 
Whye Kit Fong for access to nuScenes-lidarseg.